\documentclass[runningheads]{llncs}
\usepackage{amssymb}
\usepackage{amsmath}
\usepackage{graphicx}
\usepackage[table,pdftex]{xcolor}
\usepackage{xspace}
\usepackage{lineno}
\usepackage{subfig}

\usepackage{booktabs}

\usepackage{bbm}
\usepackage{dsfont}
\usepackage[misc]{ifsym}
\usepackage{hyperref}
\begin{document}
\title{Nesterov Accelerated ADMM for Fast Diffeomorphic Image Registration}
%
%
\author{Paper ID 1502}
 \author{Alexander Thorley\inst{1}  \and Xi Jia\inst{1} \and
 Hyung Jin Chang \inst{1} \and
 Boyang Liu \inst{2}  \and 
 Karina Bunting \inst{2} \and
 Victoria Stoll \inst{2} \and
 Antonio de Marvao \inst{3} \and 
 Declan P. O'Regan \inst{3}\and \\
 Georgios Gkoutos \inst{4} \and
 Dipak Kotecha \inst{2} \and
 Jinming Duan\inst{1}$^{\textrm{(\Letter)}}$ }
 \authorrunning{F. Author et al.}
 %
 \institute{$^1$School of Computer Science, University of Birmingham, UK\\
 \email{j.duan@cs.bham.ac.uk}\\
 $^2$Institute of Cardiovascular Sciences, University of Birmingham, UK\\
 $^3$MRC London Institute of Medical Sciences, Imperial College London, UK \\
 $^4$Institute of Cancer and Genomic Sciences, University of Birmingham, UK}



\maketitle          
\begin{abstract}
Deterministic approaches using iterative optimisation have been historically successful in diffeomorphic image registration (DiffIR). Although these approaches are highly accurate, they typically carry a significant computational burden. Recent developments in stochastic approaches based on deep learning have achieved sub-second runtimes for DiffIR with competitive registration accuracy, offering a fast alternative to conventional iterative methods. In this paper, we attempt to reduce this difference in speed whilst retaining the performance advantage of iterative approaches in DiffIR. We first propose a simple iterative scheme that functionally composes intermediate non-stationary velocity fields to handle large deformations in images whilst guaranteeing diffeomorphisms in the resultant deformation. We then propose a convex optimisation model that uses a regularisation term of arbitrary order to impose smoothness on these velocity fields and solve this model with a fast algorithm that combines Nesterov gradient descent and the alternating direction method of multipliers (ADMM). Finally, we leverage the computational power of GPU to implement this accelerated ADMM solver on a 3D cardiac MRI dataset, further reducing runtime to less than 2 seconds. In addition to producing strictly diffeomorphic deformations, our methods outperform both state-of-the-art deep learning-based and iterative DiffIR approaches in terms of dice and Hausdorff scores, with speed approaching the inference time of deep learning-based methods.

\end{abstract}

\section{Introduction}
Over the past two decades, diffeomorphic image registration (DiffIR) has become a powerful tool for deformable image registration. The goal of DiffIR is to find a smooth and invertible spatial transformation between two images, such that every point in one image has a corresponding point in the other. Such transformations are known as diffeomorphisms and are fundamental inputs for computational anatomy in medical imaging. 

Beg et als. pioneering work developed the large-deformation diffeomorphic metric mapping (LDDMM) framework \cite{beg2005computing}, formulated as a variational problem where the diffeomorphism is represented by a geodesic path (flow) parameterised by time-dependent (non-stationary) smooth velocity fields through an ordinary differential equation. Vialard et al. proposed a geodesic shooting algorithm \cite{vialard2012diffeomorphic} in which the geodesic path of diffeomorphisms can be derived from the Euler-Poincaré differential (EPDiff) equation given only the initial velocity field. In this case, geodesic shooting requires only an estimate of the initial velocity rather than the entire sequence of velocity fields required by LDDMM \cite{beg2005computing}, thus reducing the optimisation complexity. Recently, Zhang et al. developed the Fourier-approximated Lie algebras for shooting (FLASH) \cite{zhang2019fast} for DiffIR, where they improved upon the geodesic shooting algorithm \cite{vialard2012diffeomorphic,singh2013vector} by speeding up the calculation of the EPDiff equation in the Fourier space. However, solutions of these approaches \cite{beg2005computing,vialard2012diffeomorphic,zhang2019fast} reply on gradient descent, which requires many iterations to converge and is therefore very slow. To reduce the computational cost of DiffIR, Ashburner represented diffeomorphsims by stationary velocity fields (SVFs) \cite{arsigny2006log}, and proposed a fast DiffIR algorithm (DARTEL) \cite{ashburner2007fast} by composing successive diffeomorphisms using scaling and squaring. However, SVFs do not provide geodesic paths between images which may be critical to statistical shape analysis. Another approach for fast DiffIR is demons \cite{vercauteren2009}, although this is a heuristic method and does not have a clear energy function to minimise. 

An alternative approach to improve DiffIR speed is to leverage deep learning, usually adopting convolutional neural networks (CNNs) to learn diffeomorphic transformations between pairwise images in a training dataset. Registration after training is then achieved efficiently by evaluating the network on unseen image pairs with or without further (instance-specific) optimisation \cite{balakrishnan2019voxelmorph}. Deep learning in DiffIR can be either supervised or unsupervised. Yang et al. proposed Quicksilver \cite{yang2017quicksilver} which utilises a supervised encoder-decoder network as the patch-wise prediction model. Quicksilver is trained with the initial momentum\footnote{Momentum is the dual form of velocity. They are connected by a symmetric, positive semi-definite Laplacian operator as defined in Eq.\eqref{eq:solutions}.} of LDDMM as the supervision signal, which does not need to be spatially smooth and hence facilitates a fast patch-wise training strategy. Wang et al. extended FLASH \cite{zhang2019fast} to DeepFLASH \cite{wang2020deepflash} in a learning framework, which predicts the initial velocity field in the Fourier space (termed the low dimensional bandlimited space in their paper). DeepFLASH is more efficient in general as the backbone network consists of only a decoder. As Quicksilver and DeepFLASH need to be trained using the numerical solutions of LDDMM \cite{singh2013vector} as ground truth, their performance may be bounded by that of LDDMM \cite{singh2013vector}. Dalca et al. proposed diffeomorphic VoxelMorph \cite{dalca2019unsupervised} that leverages the U-Net \cite{ronneberger2015u} and the spatial transformer network \cite{jaderberg2015spatial}. Mok et al. developed a fast symmetric method (SymNet) \cite{mok2020fast} that guarantees topology preservation and invertibility of the transformation. Both Voxelmorph and SymNet impose diffeomorphisms using SVFs in an unsupervised fashion. Whilst the fast inference speed of deep learning is a clear benefit, the need for large quantities of training data and the lack of robust mathematical formulation in comparison with deterministic iterative approaches means both paradigms still have their advantages.

The goal of this study is to reduce the speed gap between learning-based and iterative approaches in DiffIR, while still retaining the performance advantage of iterative approaches. We first propose an iterative scheme to decompose a diffeomorphic transformation into the compositions of a series of small non-stationary velocity fields. Next, we define a clear, general and convex optimisation model to compute such velocity fields. The data term of the model can be either ${\cal L}^1$ or ${\cal L}^2$ norm and the ${\cal L}^2$ regularisation term uses a derivative of arbitrary order whose formulation follows the multinomial theorem. We then propose a fast solver for the proposed general model, which combines Nesterov acceleration \cite{nesterov1983method} and the alternating direction method of multipliers (ADMM) \cite{boyd2011distributed,goldstein2014fast,lu2016implementation}. By construction, all resulting subproblem solutions are point-wise and closed-form without any iteration. As such, the solver is very accurate and efficient, and can be implemented using existing deep learning frameworks (eg. PyTorch) for further acceleration via GPU. We show on a 3D cardiac MRI dataset that our methods outperform both state-of-the-art learning and optimisation-based approaches, with a speed approaching the inference time of learning-based methods.  

\section{Diffeomorphic Image Registration}
Let $\phi : \mathbf{R}^3 \to \mathbf{R}^3$ denote the deformation field that maps the coordinates from one image to those in another image. Computing a diffeomorphic deformation can be treated as modelling a dynamical system \cite{beg2005computing}, given by the ordinary differential equation (ODE): $\partial{\bf{\phi}}/\partial t  = {\bf{v}}_t({\bf{\phi}}_t)$, where ${\bf{\phi}}_0 = {\rm{Id}}$ is the identity transformation and ${\bf{v}}_t$ indicates the velocity field at time t ($\in [0,1]$). A straightforward approach to numerically compute the ODE is through Euler integration, in which case the final deformation field $\phi_1$ is calculated from the compositions of a series of small deformations. More specifically, a series of $N$ velocity fields are used to represent the time varying velocity field ${\bf{v}}_t$. For $N$ uniformly spaced time steps $(0, t_1, t_2, ..., t_{N-2}, t_{N-1} )$, $\phi_1$ can be achieved by:
\begin{equation} \label{eq:ode}
    \phi_1=\left({\rm{Id}}+\frac{{\bf{v}}_{t_{N-1}}}{N}\right)\circ \left({\rm{Id}}+\frac{{\bf{v}}_{t_{N-2}}}{N}\right)\circ \hdots \circ \left({\rm{Id}}+\frac{{\bf{v}}_{t_{1}}}{N}\right)\circ \left({\rm{Id}}+\frac{{\bf{v}}_{0}}{N}\right),
\end{equation}
where $\circ$ denotes function composition. This greedy composite approach is seen in some DiffIR works \cite{christensen1996deformable,vercauteren2009} and if the velocity fields ${\bf{v}}_{t_{i}}, \forall i\in \{0,...,N-1\}$ are sufficiently small whilst satisfying some smoothness constraint, the resulting compositions should result in a deformation that is diffeomorphic \cite{christensen1995topological}.



The fact that the numerical solver \eqref{eq:ode} requires small and smooth velocity fields motivates us to consider classical optical flow models, among which a simple yet effective method was proposed by Horn and Schunck (HS), who computed velocity $\bf{v}$ through $ \min_{\bf{v}} \{\frac{1}{2}  \| \langle \nabla I_1, {\bf{v}} \rangle + I_1 - I_0 \|^2 + \frac{\lambda}{2}  \| \nabla {\bf{v}} \|^2$\} \cite{horn1981determining}. In the minimisation problem, the first data term imposes brightness constancy, where $I_0 \in {\mathbb{R}^{MNH}}$ ($MNH$ is the the image size) and $I_1 \in {\mathbb{R}^{MNH}}$ are a pair of input images and $\langle \cdot, \cdot \rangle$ denotes inner product. The second regularisation term induces spatial smoothness on ${\bf{v}}\in ({\mathbb{R}^{MNH}})^3 $, where $\lambda$ is a hyperparameter and $\nabla$ denotes the gradient operator. As the HS data term is based on local Taylor series approximations of the image signal, it is only capable of recovering small velocity fields, which together with the smooth regularisation makes the HS model perfect to combine with \eqref{eq:ode} for effective diffeomorphic registration. 

Based on this observation, we propose a simple iterative scheme for DiffIR. For a pair of images $I_0$ (target image) and $I_1$ (source image), first we solve the HS model for ${{\bf{v}}^*_{t_{N-1}}}$. Of note, as both terms in the HS model are linear and quadratic, if we derive the first-order optimality condition with respect to ${\bf{v}}$ from the model, we end up with a system of linear equations (ie. regularised normal equation), which can be solved analytically by matrix inversion. We then warp the source image $I_1$ to $I_1^\omega$ using $I_1^\omega = I_1 \circ ({\rm{Id}}+{\bf{v}}^*_{t_{N-1}})$. Note that we set the denominator coefficient $N$ to 1 in this paper. Next, we pass $I_1^\omega$ and $I_0$ to the HS model again and solve for ${{\bf{v}}^*_{t_{N-2}}}$, with which we update the warped source image $I_1^\omega$ with $I_1^\omega = I_1 \circ ({\rm{Id}}+{\bf{v}}^*_{t_{N-1}})\circ ({\rm{Id}}+{\bf{v}}^*_{t_{N-2}})$. We repeat this iterative process until that the final warped source image $I_1^\omega = I_1 \circ ({\rm{Id}}+{\bf{v}}^*_{t_{N-1}})\circ ({\rm{Id}}+{\bf{v}}^*_{t_{N-2}})\circ \hdots \circ \left({\rm{Id}}+{{\bf{v}}^*_{t_{1}}}\right)\circ \left({\rm{Id}}+{{\bf{v}}^*_{0}}\right)$ is close to the target image $I_0$ within a pre-defined, small and positive tolerance. We note that unlike \cite{beg2005computing,vialard2012diffeomorphic,zhang2019fast,singh2013vector} there is no need for the proposed iterative scheme to pre-define the number of velocity fields as it is automatically computed based on the similarity between the warped and target images, however we can and will impose a restriction on the number of iterations to reduce computation time.

The HS model however suffers from the fact that the quadratic data term is not robust to outliers and that the first-order diffusion regularisation performs poorly against large deformations (which often requires an additional affine linear pre-registration to be successful \cite{fischer2003curvature}). To circumvent these shortcomings of the HS model, we propose a new, general variational model, given by
\begin{equation} \label{eq:OF}
\min_{\bf{v}} \left \{ \frac{1}{s} \|\rho({\bf{v}}) \|^s + \frac{\lambda}{2} \|\nabla^n {\bf{v}} \|^2 \right \},
\end{equation}
where $\rho({\bf{v}})=  \langle \nabla I_1, {\bf{v}} \rangle + I_1 - I_0 $, $s=\{1,2\}$ and $n>0$ is an integer number. $\nabla^n$ denotes an arbitrary order gradient. For a scalar-valued function $f(x,y,z)$, $\nabla^n f =  \arraycolsep=1.4pt\def\arraystretch{0.1} 
[ (\begin{array}{c}
n\\
k_1,k_2,k_3
\end{array})\frac{\partial^n f}{\partial x^{k_1}\partial y^{k_2} \partial z^{k_3} }]^{\rm{T}}$, where $k_i=0,...,n$, $i \in \{1,2,3\}$ and $k_1+k_2+k_3=n$. Moreover, $\arraycolsep=1.4pt\def\arraystretch{0.1} (\begin{array}{c}
n\\
k_1,k_2,k_3
\end{array})$ are known as multinomial coefficients which are computed by $\frac{n!}{k_1!k_2!k_3!}$. Note that if we set $s=2$ and $n=1$, the proposed model is recovered to the original HS model. However, if we set $s = 1$, the data term will be the  sum  absolute differences (SAD) \cite{zach2007duality} which is more robust to outliers in images. On the other hand, the proposed regularisation has a general form, which allows reconstruction of a smooth polynomial function of arbitrary degree for the velocity. Unfortunately, solving the proposed model becomes non-trivial due to the non-differentiality in the data term and the generality in the derivative order. We note that even if we only solve the original HS model, we still need to do matrix inversion which is impractical for 3D high-dimensional data used in this paper. To address these issues, we propose an effective algorithm in the following section to minimise the proposed convex model efficiently. 

\section{Nesterov Accelerated ADMM}
In order to minimise the convex problem \eqref{eq:OF} efficiently, we utilise the alternating direction method of multipliers (ADMM) \cite{boyd2011distributed,goldstein2014fast,lu2016implementation} accelerated by Nesterov's approach for gradient descent methods \cite{nesterov1983method}. First, we introduce an auxiliary primal variable ${\bf{w}} \in ({\mathbb{R}^{MNH}})^3 $, converting (\ref{eq:OF}) into the following equivalent form
\begin{equation} \label{eq:constrainedproblem}
\min_{\bf{v}} \left \{ \frac{1}{s} \|\rho({\bf{v}}) \|^s   + \frac{\lambda}{2} \|\nabla^n {\bf{w}} \|^2\right \} \; s.t.\;  {\bf{w}} = {\bf{v}} .
\end{equation}
The introduction of the constraint ${\bf{w}} = {\bf{v}}$ decouples ${\bf{v}}$ in the regularisation from that in the data term so that we can derive closed-form, point-wise solutions with respect to both variables regardless of the non-differentiality and generality of the problem \eqref{eq:OF}. To guarantee an optimal solution, the above constrained problem can be converted to a saddle problem solved by the proposed fast ADMM. Let ${\cal {L}}_{\cal {A}}({\bf{v}}, {\bf{w}}; {\bf{b}}) $ be the augmented Lagrange functional of \eqref{eq:constrainedproblem}, defined as 
\begin{equation} \label{eq:ALM}
{\cal {L}}_{\cal {A}}({\bf{v}}, {\bf{w}}; {\bf{b}}) = \frac{1}{s} \|\rho({\bf{v}}) \|^s   + \frac{\lambda}{2} \|\nabla^n {\bf{w}} \|^2 + \frac{\theta}{2} \|{\bf{w}}- {\bf{v}} - {\bf{b}}\|^2,
\end{equation}
where ${\bf{b}} \in ({\mathbb{R}^{MNH}})^3 $ is the augmented Lagrangian multiplier (aka. dual variable), and $\theta > 0$ is the penalty parameter which has impact on how fast the minimisation process converges. It is known that one of the saddle points for \eqref{eq:ALM} will give a minimiser for the constrained minimisation problem \eqref{eq:constrainedproblem}. We use the Nesterov accelerated ADMM to optimise variables in \eqref{eq:ALM}, which is given as  
\begin{equation} \label{eq:subproblems}
\left\{ \begin{array}{l}
{\bf{v}}^{k} = \mathop {\arg \min }\limits_{{\bf{v}}} \frac{1}{s} \|\rho({\bf{v}})\|^s + \frac{\theta}{2}  \|\widehat {\bf{w}}^k - {\bf{v}} -  \widehat{\bf{b}}^k  \|^2 \\
{\bf{w}}^{k} = \mathop {\arg \min }\limits_{\bf{w}} \frac{\lambda}{2}  \|\nabla^n {\bf{w}} \|^2 + \frac{\theta}{2}  \|{\bf{w}} - {\bf{v}}^{k} - \widehat{\bf{b}}^k\|^2 \\
{\bf{b}}^{k} =  \widehat{\bf{b}}^{k}  + {\bf{v}}^{k} -{\bf{w}}^{k} \\
\alpha^{k+1} = \frac{1 + \sqrt{1+4(\alpha^k)^2}}{2} \\
\widehat{\bf{w}}^{k + 1} = {\bf{w}}^{k} + \frac{\alpha^k - 1}{\alpha^{k+1}}({\bf{w}}^{k} - {\bf{w}}^{k - 1}) \\
\widehat{\bf{b}}^{k+1} = {\bf{b}}^{k} + \frac{\alpha^k - 1}{\alpha^{k+1}}({\bf{b}}^{k} - {\bf{b}}^{k - 1}) 
\end{array} \right..
\end{equation}
At the beginning of the iteration, we set $\widehat {\bf{w}}^{-1} = \widehat {\bf{w}}^{0} = {\bf{0}}$, $\widehat {\bf{b}}^{-1} = \widehat {\bf{b}}^{0}  = {\bf{0}}$ and $\alpha^0 = 1$. $k$ above denotes the number of iterations. As can be seen from \eqref{eq:subproblems}, we decompose \eqref{eq:ALM} into two convex subproblems with respect to ${\bf{v}}$ and ${\bf{w}}$, update the dual variable ${\bf{b}}$, and then apply Nesterov’s acceleration to both primal variable ${\bf{w}}$ and dual variable ${\bf{b}}$. This accelerated iterative process is repeated until the optimal solution ${\bf{v}}^{*}$ is found. The convergence of this accelerated ADMM has been proved in \cite{goldstein2014fast} for convex problems and the authors have also demonstrated the accelerated ADMM has a convergence rate of ${\cal O}(1/k^2)$, which is the optimal rate for first-order algorithms. 
Moreover, due to the decoupling we are able to derive a closed-form, point-wise solution for both subproblems of ${\bf{v}}$ and ${\bf{w}}$ 
\begin{equation} \label{eq:solutions}
    \begin{cases}
      {\bf{v}}^{k} = \widehat{\bf{w}}^k  - \widehat{\bf{b}}^k  - \frac{{{{\hat{z} }}}}{{\max \left( {\left| {{{\hat{z} }}} \right|,1} \right)}} \frac{\nabla I_1}{\theta} & \text{ if } s = 1\\ 
      ({\bf{J}} {\bf{J}} ^{\rm{T}} + \theta {\mathds{1}}) {\bf{v}}^{k} = \theta (\widehat{\bf{w}}^k   - \widehat{\bf{b}}^k) - {\bf{J}}(I_1 - I_0) & \text{ if } s = 2 \\
      {\bf{w}}^{k} = {{\cal{F}}^{ - 1}}( {{{\cal{F}}( {\bf{v}}^{k} + \widehat{\bf{b}}^k )}}/(\lambda {\cal{F}} ( {{\Delta}^n}) + \theta ))
    \end{cases} .
\end{equation}
In the case of $s=1$, we have $\hat{z} = \theta{\rho({ \widehat{\bf{w}}^k  - \widehat{\bf{b}}^k })}/{(|\nabla I_1|^2+ \epsilon)}$, where $\epsilon>0$ is a small, positive number to avoid division by zeros. 
In the case of $s=2$, ${\bf{J}} {\bf{J}} ^ {\rm{T}}$ (where ${\bf{J}}=\nabla I_1$) is a rank-1 outer product and ${\mathds{1}}$ is an identity matrix. As the respective $\bf{v}$-subproblem in this case is differentiable, we can derive the Sherman–Morrison formula (ie. see \eqref{eq:solutions} middle equation) by directly differentiating this subproblem. Due to the identity matrix, the Sherman–Morrison formula \cite{vercauteren2009} will lead to a closed-form, point-wise solution to ${\bf{v}}^{k} $. In the last equation, ${\cal {F}}$ and ${\cal {F}}^{-1}$ respectively denote the discrete Fourier transformation (DFT) and inverse DFT, and $\Delta^n$ is the $nth$-order Laplace operator which are discretised via the finite difference \cite{lu2016implementation,duan2016edge}. In this case, ${\cal{F}} ( {{\Delta}^n}) = 2^n(3 -  \cos (\frac{{2\pi p}}{M}) - \cos (\frac{{2\pi q}}{N}) -  \cos (\frac{{2\pi r}}{H}) )^n$, where $p\in[0,M)$, $q\in[0,N)$ and $r\in[0,H)$ are grid indices. We note that all three solutions are closed-form and point-wise, and therefore can be computed very efficiently in 3D. 



\section{Experimental Results}
\textbf{Dataset:} The dataset used in our experiments consists of 220 pairs of 3D high-resolution (HR) cardiac MRI images corresponding to the end diastolic (ED) and end systolic (ES) frames of the cardiac cycle. The raw volumes have a resolution of 1.2$\times$1.2$\times$2.0$mm^3$. HR imaging requires only one single breath-hold and therefore introduces no inter-slice shift artifacts \cite{duan2019automatic}. Class segmentation labels at ED and ES are also contained in the dataset for the right ventricle (RV), left ventricle (LV) and left ventricle myocardium (LVM). All images are resampled to 1.2$\times$1.2$\times$1.2$mm^3$ resolution and cropped or padded to matrix size 128$\times$128$\times$96. To train comparative deep learning methods and tune hyperparameters in different methods, the dataset is split into 100/20/100 corresponding to training, validation and test sets. We report final quantitative results on the test set only.

\begin{figure} 
\includegraphics[width=\textwidth]{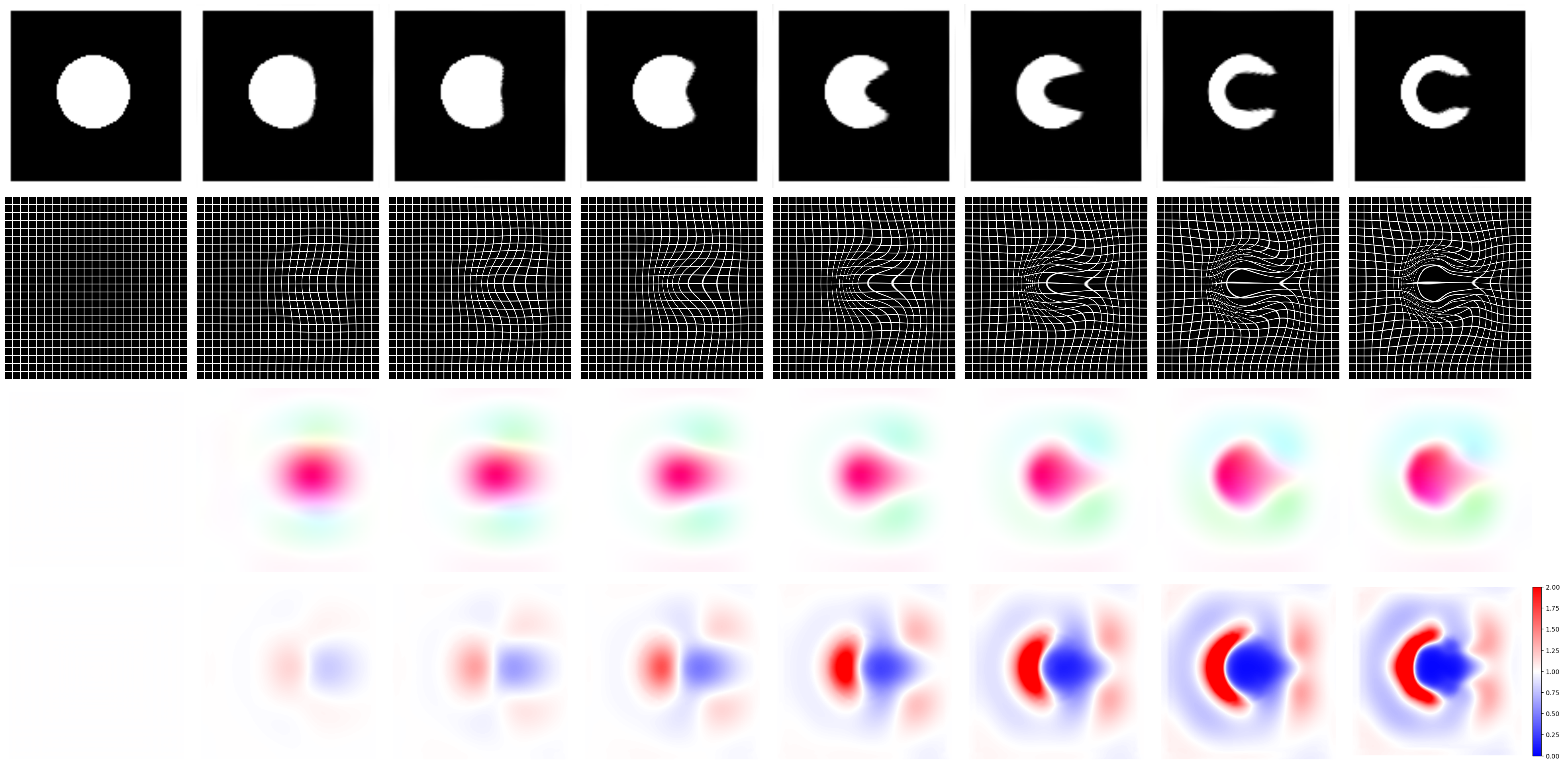} 
\vspace{-20pt}
\caption{A toy experiment showing evolution of diffeomophism throughout iterations using our method. Rows one to four correspond to image, grid deformation, HSV deformation and jacobian determinant, respectively. Columns one to eight correspond to increasing iterations.} \label{fig:box_plot}
\vspace{-20pt}
\end{figure}

\textbf{Hyperparameter tuning:} We embed our approach into a three-level pyramid scheme to avoid convergence to local minima, resulting in a total of three loops in the final implementation. We tune two versions of hyperparameters for our model, the first with limited iterations to minimise runtime, and the second without such restrictions leading to a slower runtime but higher accuracy. For the first, we restrict the iteration number within the ADMM loop in \eqref{eq:subproblems} to ten in the first pyramid level and five in the following two levels. We also cap the number of warpings in \eqref{eq:ode} to ten in the first two pyramid levels and five in the final level. For the second, iterations are only terminated when a convergence threshold (range $[0.01, 0.13]$) for the ADMM loop is met and the difference between two warpings is below $2\%$. This range of thresholds ensures runtime is competitive with other iterative methods whilst increasing accuracy compared to our first version. For both versions, the validation set is used to tune hyperparameters $\lambda$ (range $[0, 60]$) and $\theta$ (range $[0.01, 0.15]$) to maximise dice score for each combination of data term (${\cal {L}}^2$ or ${\cal {L}}^1$) and regulariser of order 1 to 3 (1\textsuperscript{st}O, 2\textsuperscript{nd}O or 3\textsuperscript{rd}O). See supplementary for the optimal values of $\lambda$ and $\theta$.

\textbf{Comparative methods:} We compare our methods against a variety of state-of-the-art iterative and deep learning-based image registration methods. For iterative methods, we use the MIRTK implementation of free form deformations (FFD) \cite{rueckert_FFD} with a three-level pyramid scheme. Control point spacing is selected to maximise dice score on the validation set. We also compare with the popular diffeomorpic demons algorithm \cite{vercauteren2009} implemented in SimpleITK \cite{SimpleITK}, using a three-level pyramid scheme. The number of iterations and smoothing $\lambda's$ are optimised on the validation set. We also compare with ANTs SyN \cite{avants_ANTS} using the official ANTs implementation with mean square distance and four registration levels. The hyper-parameters such as smoothing parameters, gradient step size, and the number of iterations are selected based on the validation set. For deep learning, we train both VoxelMorph \cite{balakrishnan2019voxelmorph} and its diffeomorphic version \cite{dalca2019unsupervised} on the training set. We adopt a lighter 5-level hierarchical U-shape network \cite{mok2020fast} than the original Voxelmorph paper \cite{balakrishnan2019voxelmorph} as our 128$\times$128$\times$96 images require significant GPU memory. 
Both VoxelMorphs are optimised under an SAD loss defining image similarity, a diffusion regularisation loss to enforce smoothness of the velocity field and a local orientation consistency loss \cite{mok2020fast} that regularises the Jacobian determinant of the velocity field. For the diffeomorphic VoxelMorph, an additional 7 scaling and squaring layers are used to encourage diffeomorphic properties in the resulting deformation. The weights of smoothness regularisation and local orientation consistency are tuned to 0.1 and 1000 respectively on the validation set, and both trained for 30000 iterations with batch size 2 and learning rate 0.0001 on a GTX 1080Ti GPU. Training takes around 13 hours. 

\begin{figure}[t]
\includegraphics[width=\textwidth]{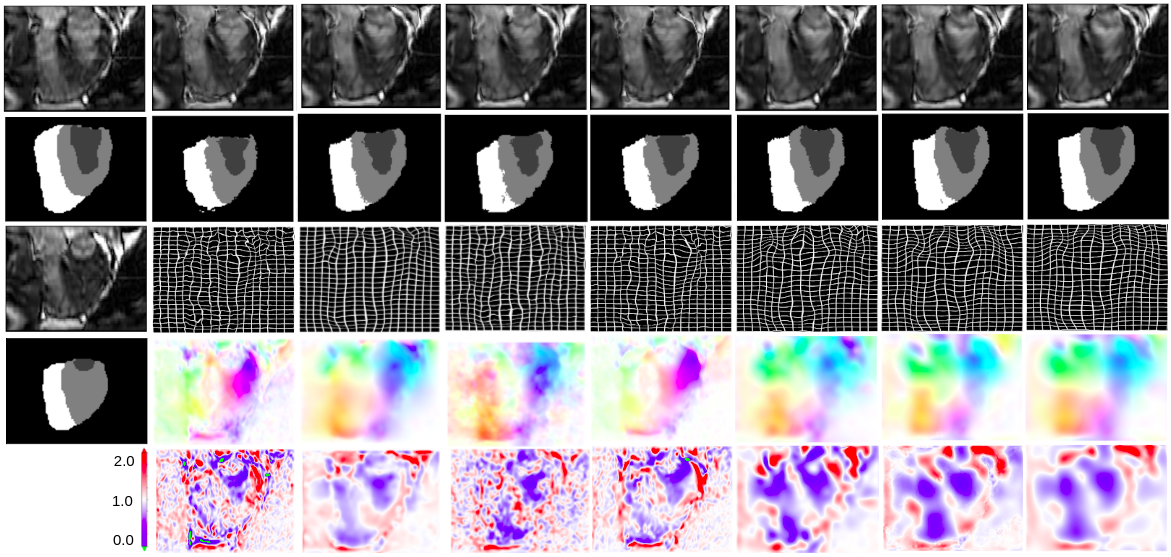}
\vspace{-20pt}
\caption{Visual results of different methods. Column 1: target image, target label, source image, source label. Columns 2-8: warped source image, warped source label, grid\textsubscript{$\phi$}, HSV\textsubscript{$\phi$}, det(J\textsubscript{$\phi$}) for (1) Voxelmorph, (2) SyN, (3) Demons, (4) diffeomorphic voxelmorph, (5) FFD, (6) ${\cal {L}}^1+$3\textsuperscript{rd}O and (7) ${\cal {L}}^2+$3\textsuperscript{rd}O, respectively.} \label{fig:full_visualisation}
\vspace{-10pt}
\end{figure}

\textbf{Quantitative results:} To evaluate performance, we use the deformations produced by each method to warp the associated source (ES) class label to its corresponding target (ED) label. The warped source and target labels are then used to compute the dice similarity coefficient and Hausdorff distance between pairs, with Hausdorff distances averaged over each 2D slice within the 3D volumes. We compute the percentage of non-positive voxels in the Jacobian J\textsubscript{$\phi$} of each deformation $\phi$ to assess its diffeomorphic properties, for which a purely diffeomorphic deformation should be zero everywhere. We also compare runtime, measuring speed of iteritive methods and inference time for two Voxelmorph models. All methods using a GPU are tested on a 16GB NVIDIA Tesla V100.

\begin{table}[]
    \vspace{-20pt}
	\centering
	\caption{Statistics of each method averaged over test set. Dice and Hausdorff scores are denoted as an average across RV, LV and LVM. Standard deviations are shown after $\pm$. Hausdorff distance is measured in $mm$ and runtime in seconds. Results of our methods using 2nd hyperparameter version are given in brackets.}
	\resizebox{1\textwidth}{!}{
		\begin{tabular}{|c|c|c|c|c|}
			\hline
			\textbf{Method} & \textbf{Dice} & \textbf{Hausdorff} & \textbf{\%} \textbf{of} $\vert \textbf{J}_{\phi}\vert\leq0$ & \textbf{Runtime} \\  \hline
			Unreg & .493$\pm$.043 & 8.404$\pm$0.894 &  -    & - 
			\\ \hline
			Voxelmorph & .709$\pm$.032 & 7.431$\pm$1.049 & 0.01$\pm$0.01 & \textbf{0.09$\pm$0.23}
			\\ \hline
			SyN & .721$\pm$.051 & 6.979$\pm$1.233 & 0.00$\pm$0.00 & 77.39$\pm$9.57
			\\ \hline
			Demons & .727$\pm$.040 & 6.740$\pm$0.998 & 0.00$\pm$0.00 & 13.01$\pm$0.17
			\\ \hline  
			Diff-Voxelmorph & .730$\pm$.031 & 6.498$\pm$0.975 & 0.01$\pm$0.01 & \textbf{0.09$\pm$0.21}
			\\ \hline
			FFD & .739$\pm$.047 & 7.044$\pm$1.200 & 0.77$\pm$0.30 & 8.98$\pm$1.97 
			\\ \hline
			${\cal {L}}^1+$1\textsuperscript{st}O & .727$\pm$.039 (.728$\pm$.040) & 6.949$\pm$1.056 (6.911$\pm$1.092) & \textbf{0$\pm$0 (0$\pm$0)} & 1.63$\pm$.26 (7.90$\pm$.79)
			\\ \hline
			${\cal {L}}^2+$1\textsuperscript{st}O & .719$\pm$.039 (.726$\pm$.041) & 7.499$\pm$1.227 (7.105$\pm$1.199) & \textbf{0$\pm$0 (0$\pm$0)} & 2.17$\pm$.50 (4.69$\pm$.34)
			\\ \hline
			${\cal {L}}^1+$2\textsuperscript{nd}O & .749$\pm$.042 (.761$\pm$.041) & 6.891$\pm$1.134 (\textbf{6.364$\pm$1.036}) & \textbf{0$\pm$0 (0$\pm$0)} & 1.43$\pm$.05 (5.02$\pm$.57)
			\\ \hline
			${\cal {L}}^2+$2\textsuperscript{nd}O & .735$\pm$.043 (.751$\pm$.045) & 6.895$\pm$1.212 (6.855$\pm$1.231) & \textbf{0$\pm$0 (0$\pm$0)} & 1.38$\pm$.04 (5.00$\pm$.41)
			\\ \hline
			${\cal {L}}^1+$3\textsuperscript{rd}O & .753$\pm$.042 (\textbf{.768$\pm$.042}) & 6.963$\pm$1.100 (6.515$\pm$1.077) & \textbf{0$\pm$0 (0$\pm$0)} & 1.57$\pm$.24 (4.49$\pm$.56)
			\\ \hline
			${\cal {L}}^2+$3\textsuperscript{rd}O & .736$\pm$.042 (.757$\pm$.046) & 7.302$\pm$1.237 (6.927$\pm$1.264) & \textbf{0$\pm$0 (0$\pm$0)} & 1.37$\pm$.03 (4.72$\pm$.41)
			\\ \hline
	\end{tabular}}\label{tbl:results_summary}
\end{table}

In Fig~\ref{fig:full_visualisation}, we show the visual comparison between all methods, where grid\textsubscript{$\phi$} shows the deformation applied to a unit grid, det(J\textsubscript{$\phi$}) the jacobian determinant and  HSV\textsubscript{$\phi$} the final composition of each iterative velocity field, with hue depicting direction and saturation magnitude. In Table~\ref{tbl:results_summary}, we show the final results of each method averaged over the test set. Results from both versions of hyperparameters are displayed with the latter in brackets. It can be seen that our accelerated ADMM outperforms all other methods by at least $1\%$ in terms of dice score despite limiting the number of iterations and warpings, increasing to nearly 3\% for our second version. There is also a clear upward tendency in dice (see Fig~\ref{fig:box_plot}), further demonstrating the superiority of our methods. Moreover, our methods consistently maintain diffeomorphic properties with no negative elements present in J\textsubscript{$\phi$}. Although the speed of our method is still slower than deep learning methods, the average runtime we achieved for our first hyperparameter version is barely over a second. 

\begin{figure}
\vspace{-10pt}
\includegraphics[width=\textwidth]{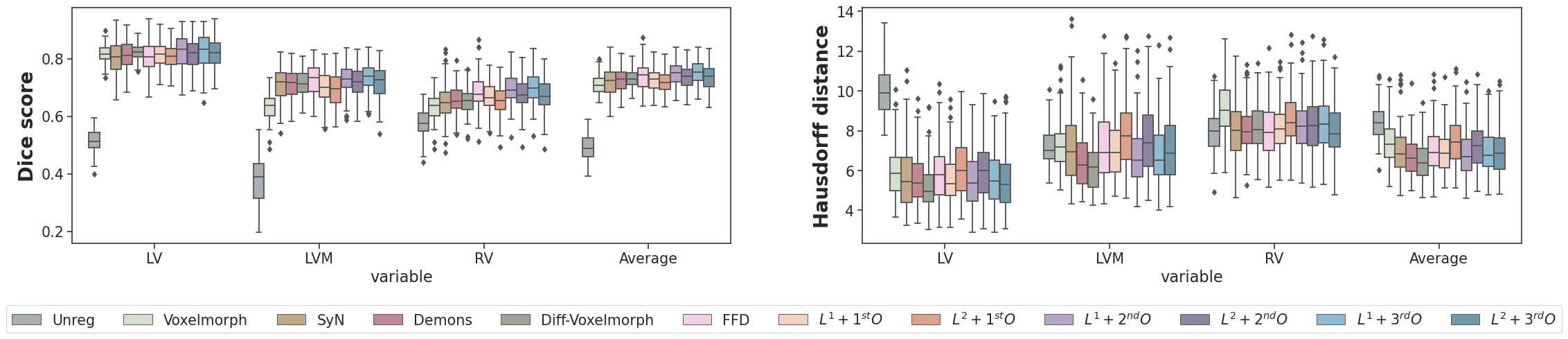}
\vspace{-18pt}
\caption{Boxplots of Dice and Hausdorff for left ventricle (LV), left ventricle myocardium (LVM) and right ventricle (RV) of all methods on test set.} \label{fig:box_plot}
\vspace{-20pt}
\end{figure}

\section{Conclusion}
In this paper, we proposed an accelerated algorithm that combines Nesterov gradient descent and ADMM, which tackles a general variational model with a regularisation term of arbitrary order. Through the compositions of a series of velocity fields produced by our accelerated ADMM, we model deformations as the solution to a dynamical system to ensure diffeomorphic properties. We implement our methods in the PyTorch framework, leveraging the power of a GPU to further accelerate our methods. In addition to reducing the difference in speed between deep learning to under 2 seconds, our methods have achieved a new state-of-the-art performance for the DiffIR task.\\ 

\noindent {\bf{Acknowledgements}}. The research is supported by the BHF Accelerator Award (AA/18/2/34218), the Ramsay Research Fund from the School of Computer Science at the University of Birmingham and the Wellcome Trust Institutional Strategic Support Fund: Digital Health Pilot Grant.


\bibliographystyle{splncs04}
\bibliography{mybibliography}





\clearpage

\title{Supplementary material}
\author{}
\institute{}
\maketitle   

\begin{figure}[h!]
\includegraphics[width=\textwidth]{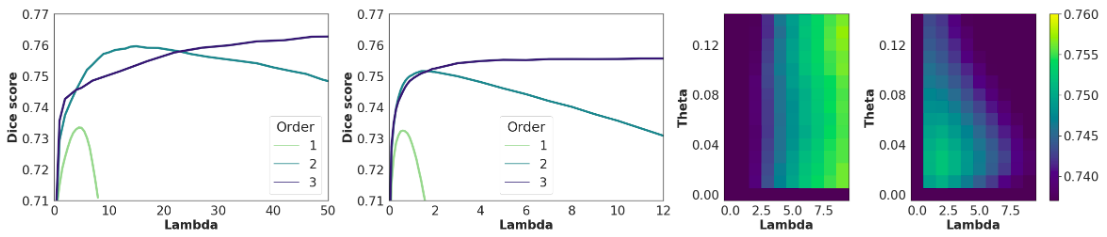}
\caption{Graph 1 and 2 show varying values of $\lambda$ and regularisation order against dice score for L1 and L2 data term respectively, with optimal $\theta$ and convergence thresholds. Graph 3 and 4 show dice score as colour for third order regularisation terms with L1 and L2 data terms respectively with varying $\theta$ and $\lambda$.} \label{fig:box_plot}
\end{figure}

\begin{figure}[h!]
\includegraphics[width=\textwidth]{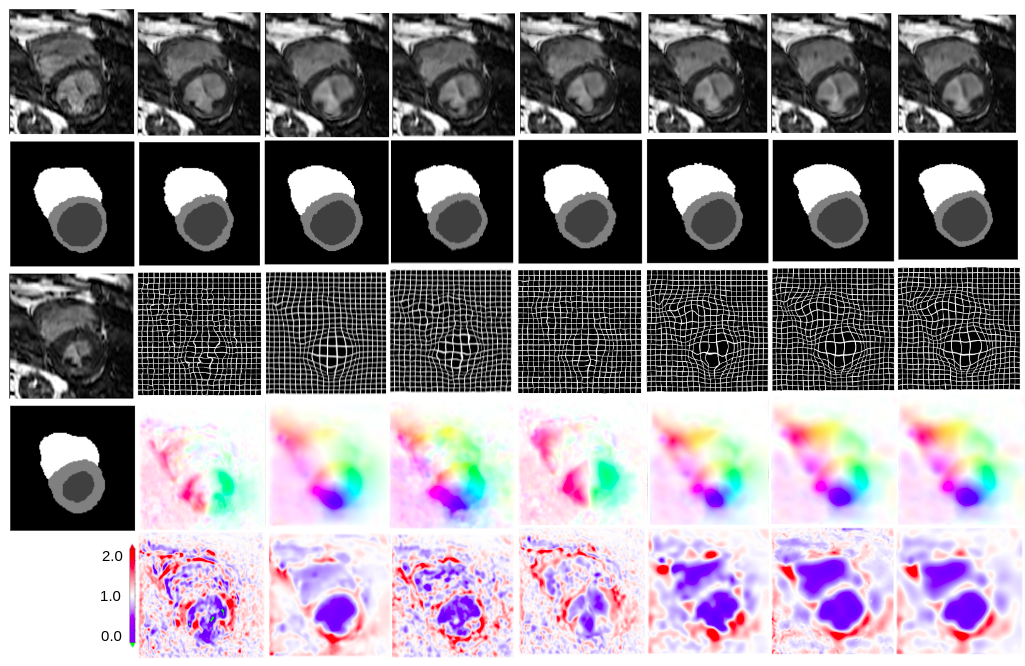}
\caption{Visual results of different methods. Column 1: target image, target label, source image, source label. Columns 2-8: warped source image, warped source label, grid\textsubscript{$\phi$}, HSV\textsubscript{$\phi$}, det(J\textsubscript{$\phi$}) for (1) Voxelmorph, (2) SyN, (3) demons, (4) diffeomorphic voxelmorph, (5) FFD, (6) ${\cal {L}}^1+$3\textsuperscript{rd}O and (7) ${\cal {L}}^2+$3\textsuperscript{rd}O, respectively.} \label{fig:full_visualisation}
\end{figure}

\end{document}